\documentclass[letterpaper, 10 pt, conference]{ieeeconf}      
\usepackage{amssymb,graphicx,amsmath,color,algorithm,algorithmic,url}
\usepackage[english]{babel}
\usepackage{blindtext}
\usepackage{afterpage}
\usepackage[noadjust]{cite} 
\usepackage{hhline,caption}
\usepackage{hyperref}
\usepackage[font=small,labelfont=bf]{caption}

\IEEEoverridecommandlockouts                              

\title{
{\bf{Educational SoftHand-A: Building an Anthropomorphic Hand\\ with Soft Synergies using LEGO\textsuperscript{\textregistered}~MINDSTORMS\textsuperscript{\textregistered}}}
}

\author{Jared K. Lepora$^{1}$, Haoran Li$^{2}$, Efi Psomopoulou$^{2}$, Nathan F. Lepora$^{2}$
\thanks{Corresponding author: n.lepora@bristol.ac.uk}
\thanks{Project website: \url{www.lepora.com/EduSoftHand-A}}
\thanks{$^{1}$Bristol Grammar School, University Road, Bristol, BS8 1SR, U.K.}
\thanks{$^{2}$School of Engineering Mathematics and Technology, and Bristol Robotics Laboratory, University of Bristol, Bristol, BS8 1UB, U.K.}
}

\begin{document}
\maketitle

\begin{abstract}
This paper introduces an anthropomorphic robot hand built entirely using LEGO MINDSTORMS: the Educational \mbox{SoftHand-A}, a tendon-driven, highly-underactuated robot hand based on the Pisa/IIT SoftHand and related hands. To be suitable for an educational context, the design is constrained to use only standard LEGO pieces with tests using common equipment available at home. The hand features dual motors driving an agonist/antagonist opposing pair of tendons on each finger, which are shown to result in reactive fine control. The finger motions are synchonized through soft synergies, implemented with a differential mechanism using clutch gears. Altogether, this design results in an anthropomorphic hand that can adaptively grasp a broad range of objects using a simple actuation and control mechanism. Since the hand can be constructed from LEGO pieces and uses state-of-the-art design concepts for robotic hands, it has the potential to educate and inspire children to learn about the frontiers of modern robotics. 
\end{abstract}


\section{INTRODUCTION}


We humans are fascinated by machines made in our own image. The human hand holds a special fascination amplified by the contradiction that it appears relatively complex in construction yet simple to use. Therefore, recreating the human hand interests adults and children, as it is practically useful for many robotics applications while enabling us to learn about ourselves and the machines we can create. Our children will form that next generation of scientists and engineers who will create new robotics technologies that will impact their lives, so it is important that we offer them fun and interesting ways to learn about robotics. 

The 2019 article ``A Century of Robotic Hands''~\cite{piazza2019century} made some key observations about past trends and future directions in
artificial hand design. One trend was towards a principled simplification of robotic hand design that maintains the advantages of an anthropomorphic structure while sensibly reducing complexity of actuation and control. A second trend was an increased adoption of soft robotic approaches, such as underactuated hands that utilize compliance or soft structures for adaptivity during grasping and manipulation. Both trends tend towards more biomimetic artificial hands, as human hands balance simplicity with functionality by making extensive use of compliance and underactuation.  

\begin{figure}[t!]
	\centering
	\begin{tabular}[b]{@{}c@{}}
        \includegraphics[width=\columnwidth,trim={5 0 5 30},clip]{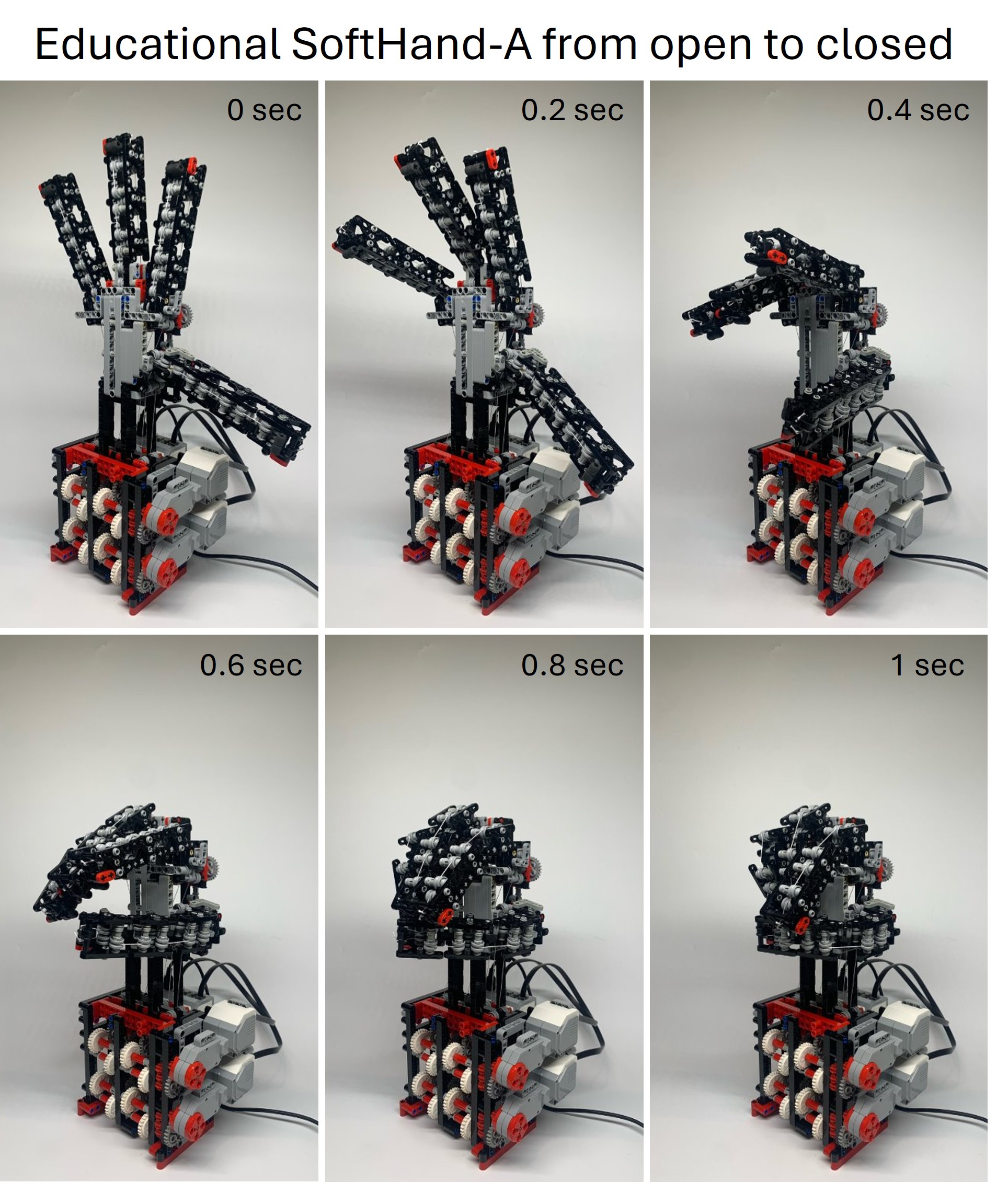} 
    \end{tabular}
	\caption{The Educational SoftHand-A is built and controlled entirely from standard LEGO\textsuperscript{\textregistered} pieces and features a highly underactuated antagonistic tendon mechanism with adaptivity from soft synergies.}
	\vspace{-.25em}
	\label{fig:1}
	\vspace{-1.25em}
\end{figure} 

Our research at Bristol Robotics Laboratory has been influenced by a robotic hand that typifies these trends: the Pisa/IIT SoftHand~\cite{catalano2014adaptive}, a highly-adaptive anthropomorphic tendon-driven hand that has naturalistic hand movements characteristic of human hand motion while using just one motor. This SoftHand has rigid phalanges connected with elastic dislocatable joints that passively reopen due to elastic ligaments. The hand adapts with contact through an adaptive synergy that synchronizes the motion of the fingers due to a common tendon looping through all the fingers. 

Inspired by the Yale OpenHand series of 3D-printed grippers~\cite{ma2017yale} and the open-sourcing of the Pisa/IIT SoftHand design~\cite{della2017quest}, we created the 3D-printable BPI (Bristol/Pisa/IIT) SoftHand~\cite{li2022brl}. For easier fabrication, the joints were simplified to pivot around linkage attachments, and soft synergies (springs) in the tendon-actuator coupling recovered some adaptivity and force transmission sacrificed by increased tendon friction due to manufacturing~\cite{li2022brl}. Then we extended the actuation to two motors in the \mbox{SoftHand-A}~\cite{li2024tactile}, with the tendons arranged as an agonist/antagonist pair on each finger that both use soft synergies. This mechanism is more biomimetic and greatly improves the hand's reactivity and force transmission by removing the need for elastic ligaments that always provide force to passively reopen the hand. 

    
    
Here, we design and build a version of the \mbox{SoftHand-A} using LEGO\textsuperscript{\textregistered} MINDSTORMS\textsuperscript{\textregistered}, so that it may serve to teach schoolchildren about the frontiers of modern robotics. The challenge is to work within the constraints of using only standard LEGO pieces and cord for tendons, yet create a state-of-the-art anthropomorphic robot hand. To do this:\\
\noindent 1) We customized the design of the 3D-printed SoftHand-A, introducing a new antagonistic tendon layout and novel soft synergy via clutch gearing to synchronize the finger motion.\\
\noindent 2) We demonstrated that this LEGO SoftHand is reactive and controllable due to its antagonistic tendon layout, using readily available tests suitable for home or school. \\
\noindent 3) Finally, we demonstrated the adaptive grasping capability of the hand using a range of object shapes and sizes.


\begin{figure}[t!]
	\centering
	\begin{tabular}[b]{@{}c@{}}
        \includegraphics[width=\columnwidth,trim={10 20 10 0},clip]{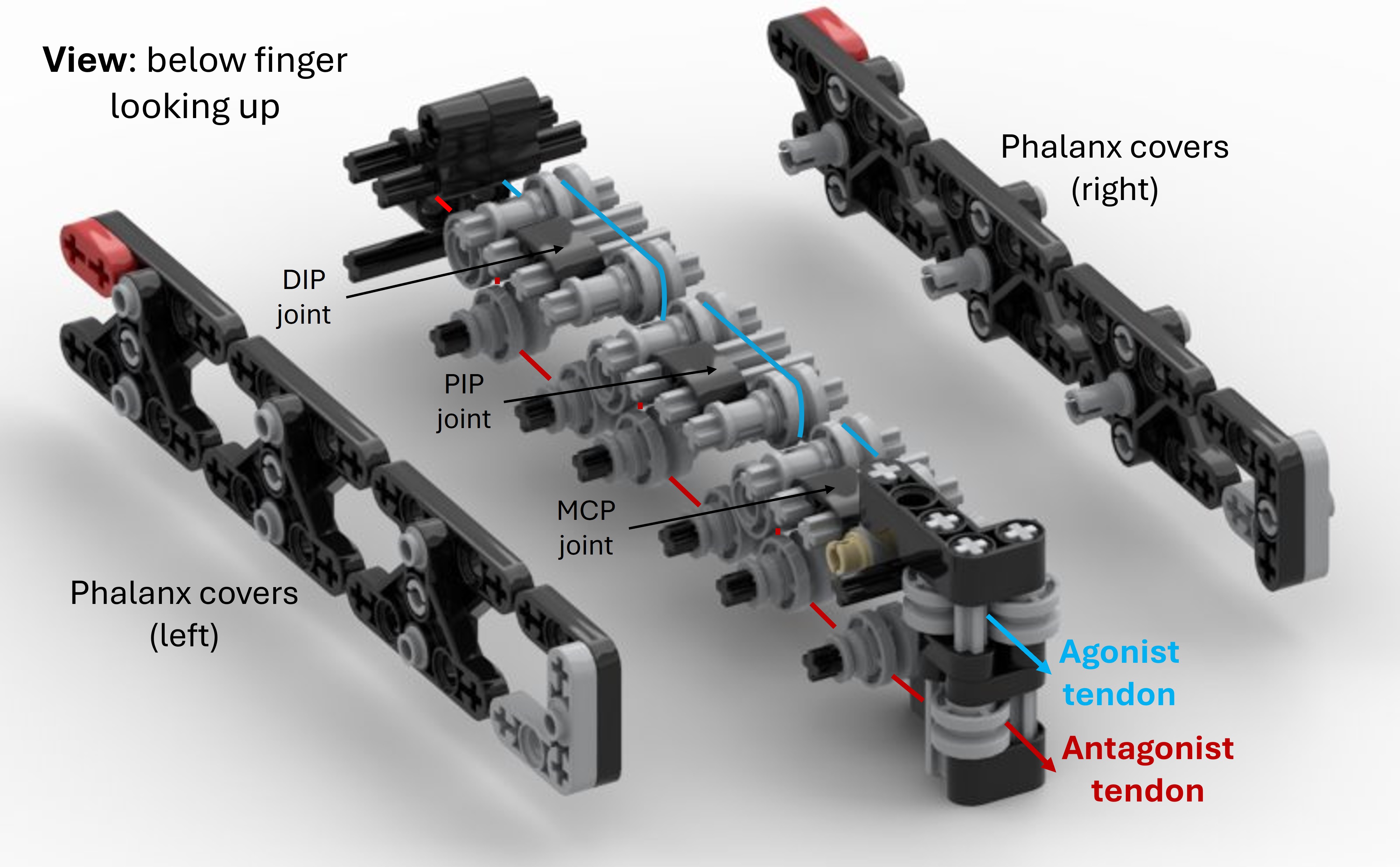} \\
        \includegraphics[width=\columnwidth,trim={5 10 5 0},clip]{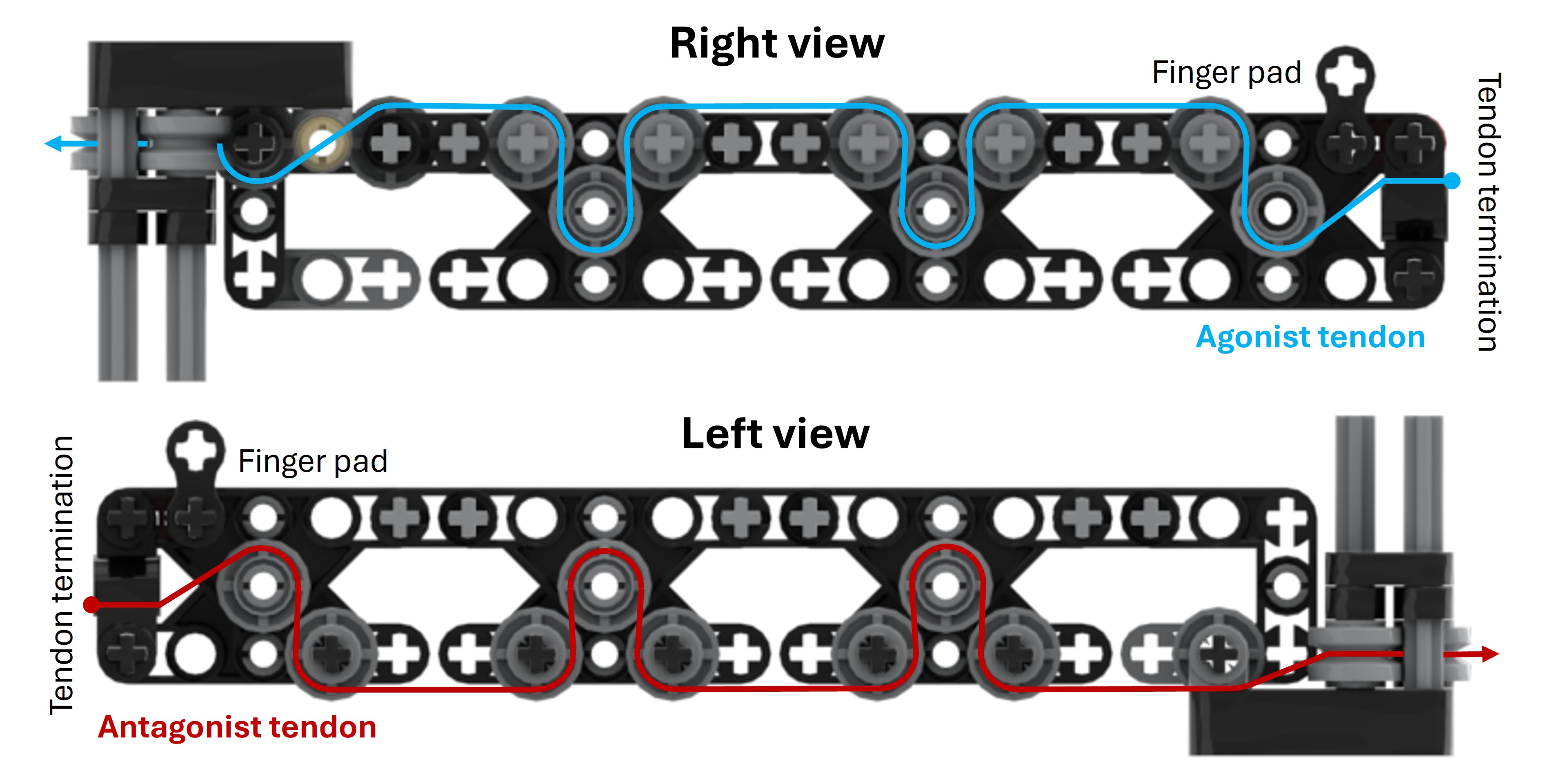} 
    \end{tabular}
	\caption{Finger design for the Educational SoftHand-A. Top: exploded view showing the axle rods that support the bearings for the driving agonist (blue) and antagonist (red) tendons. These are held in place with the phalanx covers. Bottom: side views showing the agonist and antagonist tendon routings.}
	\label{fig:2}
	\vspace{-1em}
\end{figure} 

\begin{figure*}[t!]
	\centering
	\begin{tabular}[b]{@{}c@{}}
        \includegraphics[width=2\columnwidth,trim={5 10 35 5},clip]{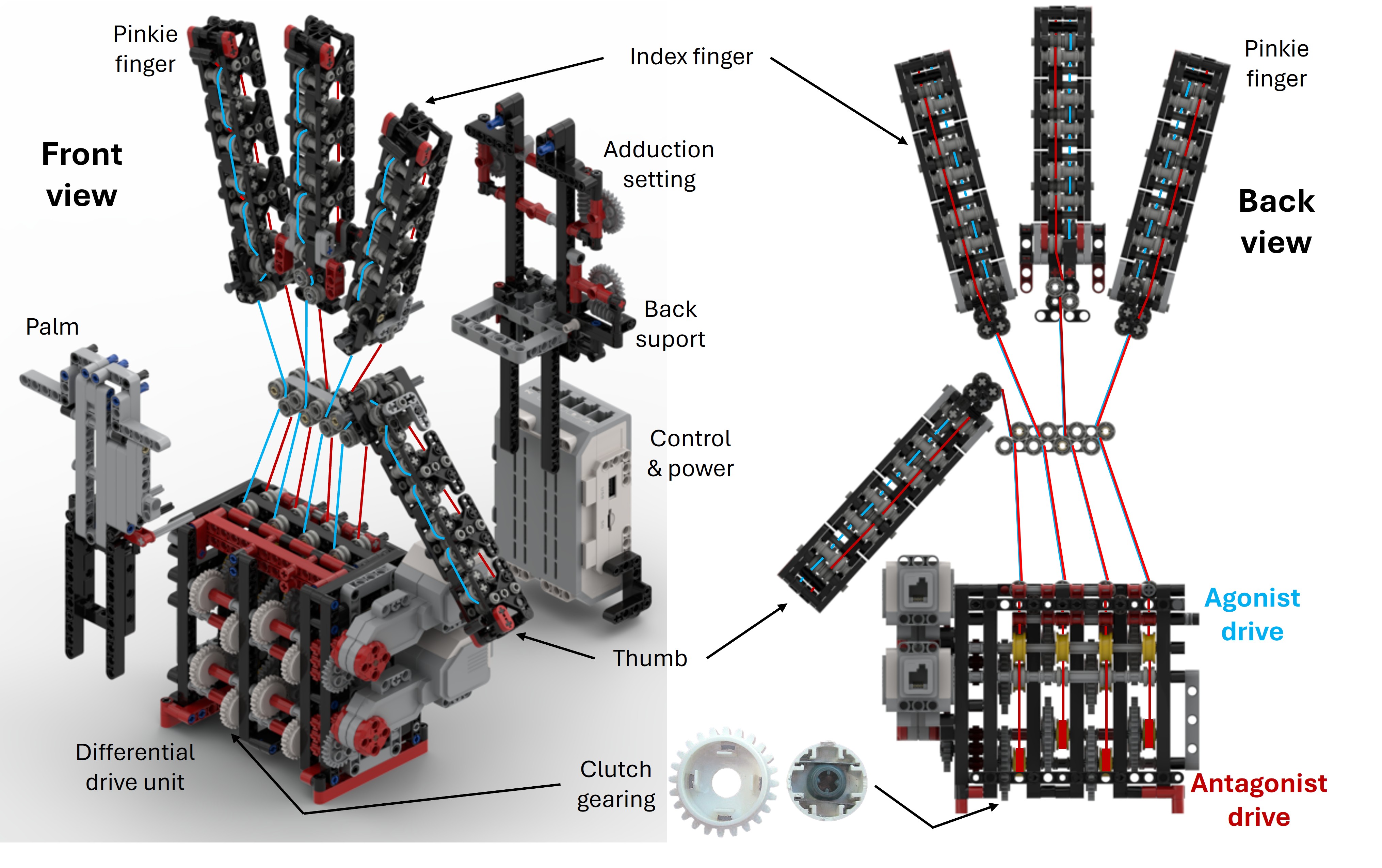} 
    \end{tabular}
	\vspace{-.5em}
	\caption{Design of the Educational SoftHand-A. Left: exploded view of the SoftHand-A main parts, including the four fingers (thumb, index, middle and pinkie). These are held by front palm and back structures, which connect to a tendon spool mechanism with two motors and programmable controller. Right: assembled front view (without the front palm and back) showing the tendon guide mechanism and feed into the tendon spool (yellow part) that uses clutch gears (colored white) for a soft synergistic action.}
	\label{fig:3}
	\vspace{-1em}
\end{figure*} 

\section{BACKGROUND AND RELATED WORK}

\subsection{Highly-underactuated Anthropomorphic Robot Hands}

The human hand is highly capable at grasping due to its mechanical structure and dexterity. The embodiment of its grasping ability is based on the synergistic operation of the ligaments, tendons, and muscles underlying the operation of human hands. The concept of soft synergy was proposed by Bicchi et al.~\cite{bicchi2011modelling}, and a transmission scheme based on this concept can address the problem of matching the actual internal force required by the controlled object with the force generated by the system. However, its mechanical structure is difficult to realize in robotic hands. Catalano et al.~\cite{catalano2014adaptive} proposed the concept of adaptive synergy and designed the Pisa/IIT SoftHand based on this principle as a highly underactuated anthropomorphic robot hand. 

The Pisa/IIT SoftHand~\cite{catalano2014adaptive} and related versions~\cite{li2022brl,li2024tactile} have anthropomorphic morphologies that can achieve naturalistic grasping motions with just one or two actuators by using tendon arrangements that synchronize joint motions. These underactuated anthorpomorphic hands benefit from novel differential mechanisms to achieve naturalistic human-like hand motions, including the parallel slider mechanism~\cite{sun2021design}, moving-pulley mechanism~\cite{4543295,gao2021anthropomorphic,chen2014mechanical} and spring groups~\cite{li2022brl,li2024tactile}. Together, these implement soft and adaptive synergies of hand motion~\cite{catalano2014adaptive,6663692,bicchi2011modelling,fan2018research}. 

\subsection{LEGO MINDSTORMS in Education}

Since their introduction in September 1998, LEGO MINDSTORMS kits have been widely used for school, college, and university education. Several factors have led to this widespread use, such as their relatively low cost (around \$200 originally), variety of connectable sensors, motors, building pieces and a programmable controller, and student interest from playing with LEGO building blocks as children~\cite{klassner_lego_2003}. LEGO robotics kits were improved in 2006 with the release of the MINDSTORMS NXT programmable sets, and upgraded again in 2013 with the MINDSTORMS EV3 used in this work. Many studies have shown the benefits of MINDSTORMS in education, particularly in teaching physics and engineering~\cite{chambers_developing_2008,danahy_lego-based_2014,gomez-de-gabriel_using_2011}. Although some LEGO robot hands have been built by hobbyists, we know of none assessed with academic peer review and none that contribute state-of-the-art designs to modern robotics research.


\section{Development of the Educational SoftHand-A}\label{Development}

\subsection{Educational SoftHand-A Design Concept}

     
The Educational SoftHand-A is an easy-to-assemble anthropomorphic robot hand inspired by the family of highly-underactuated robotic hands that originated with the Pisa/IIT SoftHand~\cite{catalano2014adaptive}. Here, the design changes are to facilitate easy assembly using LEGO building blocks. 

Overall, the Educational SoftHand-A has 4 identical fingers arranged as the index, middle, and pinkie fingers and an opposing thumb, with two opposing groups of tendons that can open and close the hand. The decision to have a 4-fingered rather than 5-fingered anthrophomorphic hand was to simplify the build and give a better-looking hand design, without greatly affecting the hand's function.

Each finger has 3 rotatory joints, giving 12 degrees for the entire hand with 2 degrees of actuation. The tendon groups are coupled to a differential driving mechanism for each actuator. This means that the hand benefits from a synergistic action in which each finger can move independently under contact with objects and the environment while all fingers are driven together to synchronize a naturalistic grasping motion of the entire hand. Together, this design and drive impart the hand with the adaptivity and grasping performance that typify the Pisa/IIT SoftHand and its derivatives. 

The Educational SoftHand-A is built entirely from standard pieces found in LEGO sets. These interconnecting components include beam connectors, gears, axle rods, pins, and bearings for guiding tendons. The hand actuation and control utilize two motors (MINDSTORMS EV3 Large Servo Motor 45502) and a programmable brick as a controller (part number 95646) from the LEGO MINDSTORMS range of educational kits for building programmable robots. 

An important standard component was LEGO bearings that attach to axle rods. The Pisa/IIT SoftHand and related versions rely on sophisticated arrangements of low-friction metal bearings to guide tendons. Likewise, the Educational SoftHand-A used over 100 plastic bearings as guides.

The design process and the illustrations used LEGO Bricklink Studio V3, which aids in the design of LEGO sets (\href{https://www.bricklink.com/v3/studio/download.page}{www.bricklink.com/v3/studio}). This professional software allows an easy exchange of the assembly instructions for others to replicate and build upon with their own designs.
 
\subsection{Finger, Phalanx and Joint Design}

\subsubsection{Modular Finger Design}

All fingers have the same design with an identical structure of four modular phalanges, to ease the design and construction process. The finger length is 145\,mm measured from its base phalange to the fingertip of the distal phalange, and its width is 30\,mm (Fig.~\ref{fig:2}). The agonist (driving) and antagonistic tendons run up each side of the finger guided by a 20 bearings (10 per side).

The four modular phalanges are of the same design, differing only in the terminating axle rod/beam assemblies at the fingertip and the bearing layout in the base to route the tendon into the palm. These include a distal fingertip phalanx, two medial phalanges, and a base phalanx that connects to the palm of the hand. Together, each finger has 22 axle rods with 20 carrying bearings for the two tendons; the most distal pair of axle beams carry short holed beams to terminate (tie) the tendons. Each phalanx is held together by a pair of phalanx covers (left and right structures, Fig.\ref{fig:2} top). 

Each finger has 3 rotatory degrees of freedom corresponding to the base metacarpophalangeal (MCP), proximal interphalangeal (PIP), and distal interphalangeal (DIP) joints. To simplify the design and control of the hand, the carpometacarpal (CMC) joint of the thumb is replaced by a standard MCP joint. In all fingers, the base, MCP, PIP and DIP phalanges are connected by phalanx connecting beams held by axle rods on the tops of each two neighboring phalanges (black parts, central structure of Fig.~\ref{fig:2}, top), which constrain the joint motion to be purely rotational around the axles. The more distal phalanx at each joint has a stop on the bottom beam below the joint beam to limit hyperextension. 

\begin{table*}[t!]
\resizebox{\textwidth}{!}{%
	\renewcommand{\arraystretch}{1}
	\centering
	\begin{tabular}{@{}lcc@{}}	
		Test & Educational SoftHand-A & 3D-printed SoftHand-A  \\
		\hline
		A1.\hspace{3em} Single finger response time closing from full extension to full flexion & 0.84\,sec & 0.39\,sec \\
        A2.\hspace{3em} Single finger response time opening from full flexion to full extension & 0.97\,sec & 0.46\,sec \\
        B1.\hspace{3em} Single finger bearing capacity & 5\,N load & 8\,N load \\
        B2.\hspace{3em} Single finger pushing capacity & 6\,N weight & 7\,N weight \\
        B3.\hspace{3em} Single finger closing force & 1.8\,N & 2\,N \\
        C1.\hspace{3em} Hand closure response time from full extension to full flexion & 0.98\,sec & 0.46\,sec\\
        C2.\hspace{3em} Hand opening response time from full flexion to full extension & 1.12\,sec & 0.59\,sec\\
		\hline
	\end{tabular}}
	\caption{Comparative single finger and whole hand test results between the Educational SoftHand-A and 3D-printed SoftHand-A~\cite{li2024tactile}.}
    \vspace{-1.5em}
	\label{tab:1}
\end{table*}

\subsubsection{Medial Phalanx Design} 

Three grooved bearings on each side of the medial phalanges guide the two tendons (Fig.~\ref{fig:2}, middle), based on the innovative tendon layout of the Pisa/IIT SoftHand~\cite{catalano2014adaptive} and its 3D-printed versions~\cite{li2022brl,li2024tactile}. The bearings are arranged in an upward isosceles triangle on the right side of the middle phalanx to facilitate flexion movement by providing a driving force to the driving tendon. Conversely, the bearings on the left side form an inverted isosceles triangle arrangement to support extension movement. Therefore, each medial phalanx has six axle rods paired with bearings to guide the two tendons. 


\subsubsection{Distal Phalanx Design}

Compared to the medial phalanges, the distal phalanx only has one pair of bearings to guide the tendon after the joint to a pair of terminating holed beams attached to the most distal top and bottom axles at the fingertips. The central pair of axles/bearings is absent to not impede the tendons (Fig.~\ref{fig:2}, top view). A rubber beam attached to the front of the fingertip acts as a finger pad to increase contact friction with grasped objects.
	
\subsubsection{Base Phalanx Design}
The base phalanx design is used both to fix the modular fingers to the palm and to guide how the two tendons enter and leave the finger from the palm (see left of Fig.~\ref{fig:2}). The phalanx's design is identical to the medial phalanges except that the top and bottom proximal axles at the finger base have fixed connecting rods that hold the base phalanx immobile relative to the palm.

\subsection{Palm Design and Tendon Layout}\label{Layout}

As the hand is tendon driven, the differential drive mechanism and motors are placed below the wrist in a separate base structure below the robot hand. Hence, the palm and four fingers are purely mechanical, in accordance with the form factor and structure of a LEGO educational robot hand. 
	
Together, eight tendons pass through the palm and four fingers, each beginning from a spool in the differential drive unit and each terminating at a beam in the fingertips (see Fig.~\ref{fig:3}). All tendons are fixed in place by knotting onto the spool at one end and terminating beam at the other. The tendons separate into two groups: an agonist group of four tendons that pass down the front of the palm and left side of the fingers (coloured blue in Figs~\ref{fig:2},\ref{fig:3}) and an antagonist group of 4 tendons that pass down the back of the palm and right side of the fingers (coloured red in Figs~\ref{fig:2},\ref{fig:3}). Each group of tendons is driven by a single actuator: an agonist servomotor to close the hand and an antagonist servomotor to open it, utilizing the soft synergies of a differential drive for an adaptive synchronized whole-hand grasping motion.  

The palm is designed to have three functions (Fig.~\ref{fig:3}): (1)~to hold the 4 fingers in an anthropomorphic arrangement with an index, middle, and pinkie finger and an opposing thumb; (2) to hold bearings to guide the two tendons from each finger down through and out of the base of the palm; and (3) to give support when grasping objects with the fingers. These functions were achieved with beams across the palm top to hold 3 fingers and their tendon bearings, and beams down one side to hold the thumb and its tendon bearings.

\subsection{Differential Drive Unit}

The differential drive distributes the torque for each motor over four tendon spools, using a common axle and gearing mechanism. The 4 agonist tendon spools are placed across the front of the drive unit, and the 4 antagonist tendon spools are arranged similarly at the back of the unit. 

Use of an innovative clutch gear mechanism to drive the spools allows the fingers to move independently under contact with objects and the environment. The clutch gears permit a maximum torque of 5\,Ncm before they disengage, 

The two actuators are MINDSTORMS EV3 Large Servo Motors (part number 45502) that connect to a programmable brick (part number 95646) from which they can be controlled. Their output torques are up to 40\,Ncm, and the motors are integrated with encoders to measure rotation. 

\begin{figure}[b!]
	\centering
	\begin{tabular}[b]{@{}c@{}}
        \includegraphics[width=0.85\columnwidth,trim={0 0 0 0},clip]{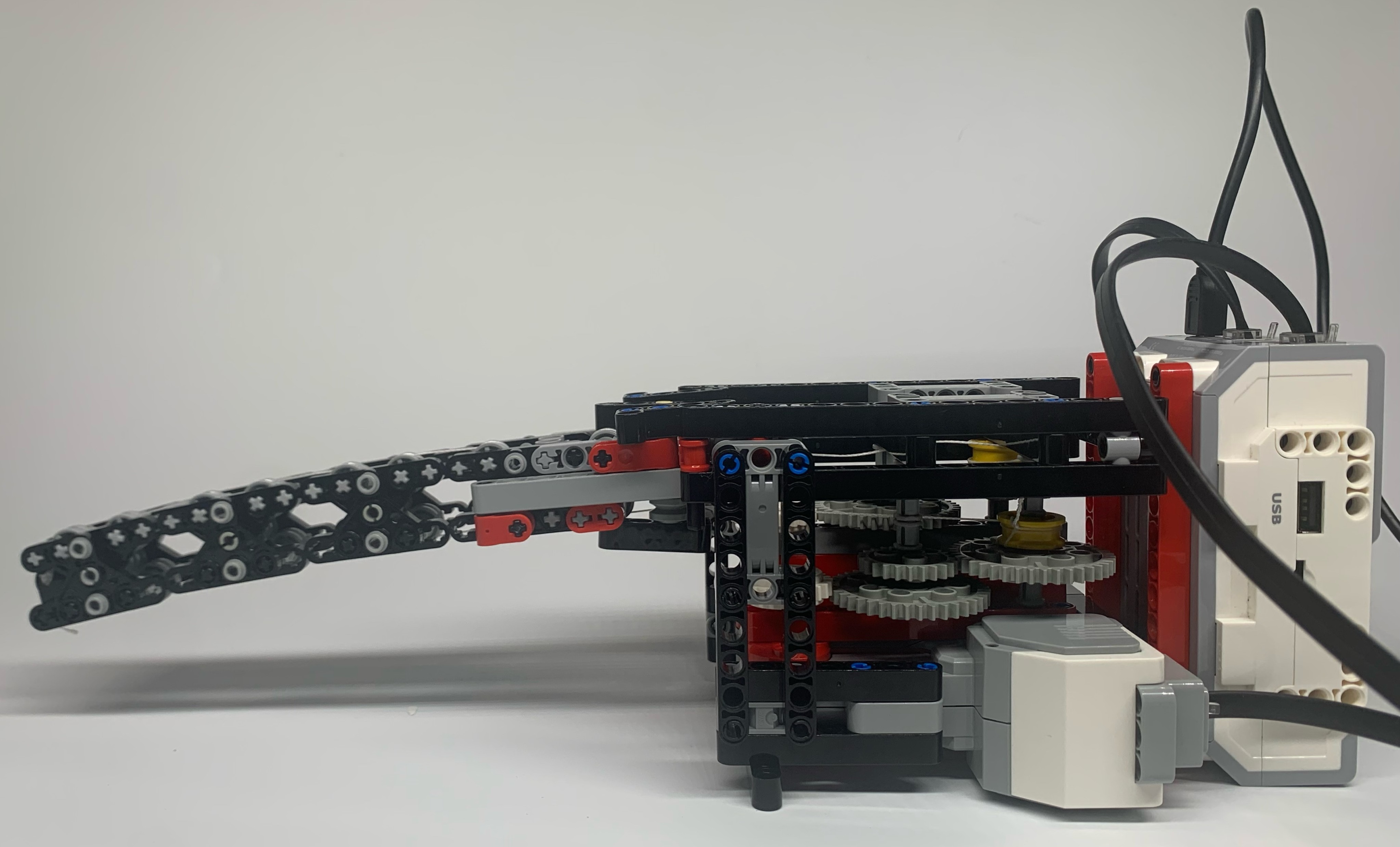} 
    \end{tabular}
	\caption{Single finger test apparatus. A single finger is mounted on a custom base containing a pair of motors that each connect to a tendon spool via a clutch gear. The apparatus enables single finger tests such as bearing capacity, pushing capacity, and closing force. }
	\label{fig:4}
\end{figure} 


\section{Results}\label{Results}

\begin{figure*}[t!]
	\centering
	\begin{tabular}[b]{@{}c@{}}
        \includegraphics[width=2\columnwidth,trim={3 0 17 5},clip]{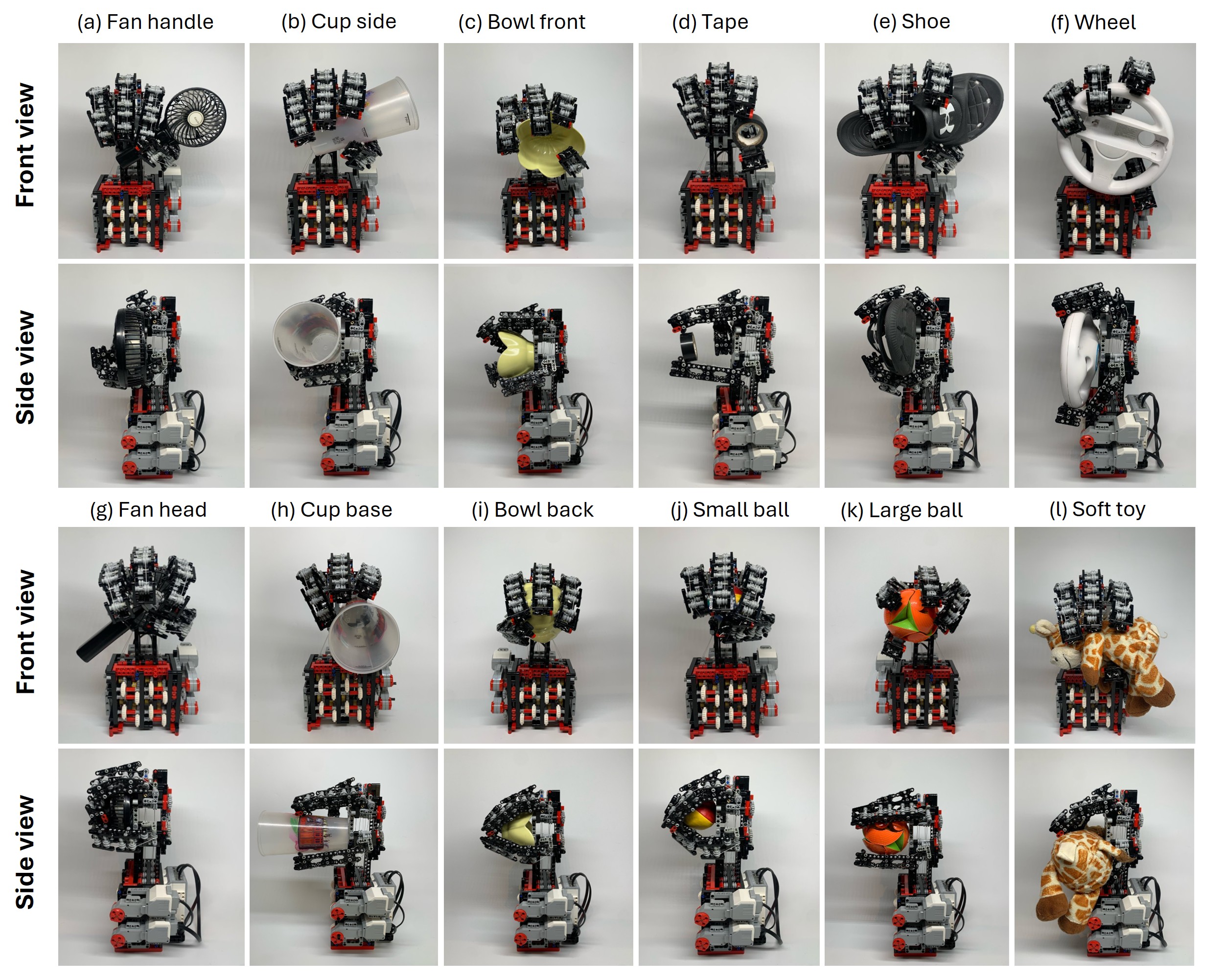} 
    \end{tabular}
	\vspace{-.5em}
	\caption{Adaptive grasping with the Educational SoftHand-A. Front and side views of each grasp are shown, using 9 distinct objects of which 3 are shown with two different grasps. The fingers have adapted to different postures depending on the shape of the grasped object.}
	\label{fig:5}
\end{figure*} 

\subsection{Finger and Hand Performance Testing}
\subsubsection{Performance Assessement Procedure}

This set of experiments examines the dynamic capabilities of a single finger and the entire SoftHand. We consider readily measurable factors that can be assessed in an educational setting:

A. Single-finger response time tests: The response time of the finger during flexion and extension is evaluated between fully open or closed finger positions. The experiments are videoed with a timer to extract response times. 

B1. Bearing capacity test: The finger’s load-bearing capability is evaluated under static conditions. Weights are applied to the medial phalange until there is a drop greater than 10\,mm of the fingertip.

B2. Pushing capacity test: The ability of the entire finger to push a stationary obstacle is assessed by finding the maximum weight moved by extending the finger. 

C. Full-hand response time tests: The response time of the educational SoftHand-A during flexion and extension is evaluated between the fully open or closed finger positions. Again, the experiments are recorded with a timer.   

\subsubsection{Performance Assessment Results}

For single-finger tests, the experimental setup is shown in Fig.~\ref{fig:4} with one finger of the SoftHand-A mounted on an agonist/antagonist drive with LEGO controller brick. For whole-hand tests, the hand shown in Fig.~\ref{fig:1} was used. The results of the single-finger and whole-hand tests are summarized in Table~\ref{tab:1}. 

For comparison, results for the SoftHand-A~\cite{li2024tactile} are also shown. As covered earlier in this paper, the SoftHand-A has a similar agonist/antagonist mechanism with soft synergies but benefits from high quality motors and a 3D-printed build with metal bearings.

A single finger of the Educational SoftHand-A is able to flex and extend in a reasonable time, taking about 1\,sec between fully open and closed positions compared to $\lesssim$~0.5\,sec for the 3D-printed SoftHand-A.

The single-finger bearing capacity and pushing capability of the Educational SoftHand-A are around 5\,--\,6\,N, a few N less than that of the 3D-printed SoftHand-A (6\,--\,8\,N). Its transmission has benefited from the lack of passive elastic components due to the antagonistic tendon mechanism, giving high force for a relatively low-power motor.

In general, the Educational SoftHand-A is able to flex and extend similarly to the speed of a single finger, taking around 1\,sec between the fully open and closed positions. Again, the speed is about half that of the 3D-printed version with a high-quality motor. Overall, the performance is still good and completely sufficient for use as an educational robot hand.  
    
\subsection{Grasping Adaptivity of the Educational SoftHand-A}
\subsubsection{Grasping Experiment Procedure}
	
To evaluate the grasping adaptivity of the Educational SoftHand-A, we mount the hand in a vertical position to test its ability to securely hold a range of household items against gravity. We placed these items in the palm of the SoftHand and drove the agonist motor to close the hand while the antagonist motor loosens its tendons to synchronize motion and maintain low stiffness of the fingers. The test items included a fan, plastic cup, bowl, tape, shoe, wheel, small/large ball, and soft toy (with weights 0.1\,--\,0.8\,kg).

\subsubsection{Grasping Experiment Results}

The Educational SoftHand-A appears to exhibit good adaptivity in grasping many different objects, as seen in images of the enclosed hand around 9 different objects (Fig.~\ref{fig:5}). This includes being able to grasp 3 items in two distinct ways (Fig.~\ref{fig:5}, columns 1\,--\,3), such as the fan from its handle or head, the cup from its side or base, and the bowl from the back or front. 

The SoftHand-A's adaptivity is evident from comparing different grasps. Comparing the bowl (column 3) with the spool (column 4), we see that the principal difference is the degree of hand closure due to both objects having a similar circular shape, but the bowl being much deeper. Meanwhile, for grasping the cup from its base (column 2, bottom), only the pinkie, middle, and thumb are in contact with the index fingers closing down the side of the object. Generally, the configurations of the fingers differ from grasp to grasp, consistent with the synergistic operation inherited from the Pisa/IIT SoftHand~\cite{catalano2014adaptive} and related hands. 

\section{Discussion}
We presented a novel underactuated and tendon-driven anthropomorphic SoftHand built entirely using standard LEGO bricks, called the Educational SoftHand-A. The design of its phalanges, palm, tendon spooling and differential drive is based on the 3D-printed \mbox{SoftHand-A}~\cite{li2024tactile}, derived from the Pisa/IIT SoftHand~\cite{catalano2014adaptive}.~The hand uses mechanical principles of soft synergies and underactuation with an antagonistic tendon arrangement. Although our \mbox{SoftHand} was made entirely using LEGO pieces, its reactivity, force transmission, and adaptive grasping compared well with other SoftHand versions, verified with single finger and entire hand tests.
    
Constraining the robotic hand to only use standard LEGO pieces led to several compromises. First, a LEGO construction gives a bulkier build than 3D-printing, so we reduced the number of digits to 3 fingers and one thumb. This design choice led to a better looking hand and did not seem to impair its functionality. Second, the 3D-printed SoftHand-A uses about 150 metal bearings to route the tendons, which were replaced with over 100 plastic LEGO bearings. Although it was fortunate that such pieces exist, they do result in higher friction on the tendons. However, this friction does not seem to result in less adaptivity of the hand during grasping (Figure~\ref{fig:5}). Third, the springs for the soft synergies in the 3D-printed SoftHand were replaced with clutch gears, which maintained adaptive grasping by synchronizing the hand motion. They also led to good force transmission and high bearing capacity. However, we found an issue that a lack of restoring tension on the tendons could lead to slack that lowered the reaction speed of the hand.   


LEGO bricks are the most popular children's toy of all time. Therefore, constructing the Educational \mbox{SoftHand-A} has the potential to be an engaging and interesting project for children to learn about modern robotics. Given that LEGO construction lends itself to trying new ideas, we expect this to lead to ingenious designs created by children. In the future, it would be exciting to see LEGO versions pushing forward the state-of-the-art research in robot dexterity. Learning about robotic hands through building them with our own hands seems a particularly apt way to engage with this technology.    
%
    


\bibliographystyle{unsrt}
\bibliography{IEEEabrv,references}

\end{document}